\documentclass[9pt,technote]{IEEEtran}

\usepackage{multicol}
\usepackage[bookmarks=true]{hyperref}
\usepackage{CJKutf8}
\usepackage{color}
\usepackage{enumerate}
\usepackage[ruled, linesnumbered]{algorithm2e}
\usepackage{amssymb, amsmath,mathtools}
\usepackage{stmaryrd}
\usepackage{graphicx}
\usepackage{setspace}
\usepackage{mathrsfs}
\usepackage{bm}
\usepackage[english]{babel}
\usepackage{blindtext}
\usepackage{epstopdf}

\newcommand{\bfalpha}{\boldsymbol{\alpha}}
\newcommand{\bfbeta}{\boldsymbol{\beta}}

\newcommand{\bfdelta}{\boldsymbol{\delta}}

\newcommand{\bfomega}{\boldsymbol{\omega}}

\newcommand{\bfepsilon}{\boldsymbol{\epsilon}}
\newcommand{\bfxi}{\boldsymbol{\xi}}

\newcommand{\bfJac}{\bm{\mathcal{J}}}

\newcommand{\diag}{\mathrm{diag}}

\newcommand{\bfb}{\bm{b}}
\newcommand{\bfc}{\bm{c}}

\newcommand{\bfe}{\bm{e}}
\newcommand{\bff}{\bm{f}}

\newcommand{\bfo}{\bm{o}}

\newcommand{\bfq}{\bm{q}}

\newcommand{\bft}{\bm{t}}

\newcommand{\bfP}{\bm{P}}
\newcommand{\bfV}{\bm{V}}

\newcommand{\bbI}{\mathbb{I}}

\newcommand{\bbR}{\mathbb{R}}

\newcommand{\calA}{{\cal A}}
\newcommand{\calB}{{\cal B}}

\def\ie{\emph{i.e.}}
\def\eg{\emph{e.g.}}

\def\etal{\emph{et al.}}

\newcommand{\tf}[3]{\prescript{}{#1}{#2}^{#3}}
\newcommand{\bmT}{\bm{T}}
\newcommand{\bmR}{\bm{R}}
\newcommand{\bmX}{\bm{X}}
\newcommand{\bmY}{\bm{Y}}
\newcommand{\bmA}{\bm{A}}
\newcommand{\bmB}{\bm{B}}
\newcommand{\bmM}{\bm{M}}
\newcommand{\bmN}{\bm{N}}

\newcommand{\bmJ}{\bm{J}}
\newcommand{\bmU}{\bm{U}}
\newcommand{\bmW}{\bm{W}}
\newcommand{\bmZ}{\bm{Z}}

\newcommand{\SE}{\bm{S}\bm{E}(3)}
\newcommand{\SO}{\bm{S}\bm{O}(3)}
\newcommand{\se}{\bm{s}\bm{e}(3)}
\newcommand{\so}{\bm{s}\bm{o}(3)}
\newcommand{\bfSigma}{{\bf{\Sigma{}}}}

\newcommand{\llangle}{\langle\langle}
\newcommand{\rrangle}{\rangle\rangle}

\begin{document}
\title{On the covariance of $\bmX$ in $\bmA\bmX = \bmX\bmB$}
\author{Huy Nguyen and Quang-Cuong Pham\thanks{The authors are with
    the School of Mechanical
  and Aerospace Engineering, Nanyang Technological University,
  Singapore 639798. Corresponding author: Huy Nguyen (email: huy.nguyendinh09@gmail.com).} }
\maketitle
\begin{abstract}
  Hand-eye calibration, which consists in identifying the rigid-body
  transformation between a camera mounted on the robot end-effector
  and the end-effector itself, is a fundamental problem in robot
  vision. Mathematically, this problem can be formulated as: solve for
  $\bmX$ in $\bmA\bmX = \bmX\bmB$. In this paper, we provide a
  rigorous derivation of the covariance of the solution $\bmX$, when
  $\bmA$ and $\bmB$ are randomly perturbed matrices. This fine-grained
  information is critical for applications that require a high degree
  of perception precision. Our approach consists in applying
  covariance propagation methods in SE(3). Experiments involving
  synthetic and real calibration data confirm that our approach can
  predict the covariance of the hand-eye transformation with
  excellent precision.
\end{abstract}


\section{Introduction}
Hand-eye calibration, which consists in identifying the rigid-body
transformation between a camera (eye) mounted on the robot
end-effector and the end-effector (hand) itself,
is a fundamental
problem in robot vision. Mathematically, this problem can be
formulated as: solve for $\bmX$ in $\bmA\bmX = \bmX\bmB$, where $\bmX$
is the unknown $4\times 4$ hand-eye transformation matrix and $\bmA$
and $\bmB$ are known $4\times 4$ transformation matrices (see details
in Section~\ref{sec:hand-eye}). Starting from the late 1980's, a large
amount of literature has been devoted to this problem, and a number
of efficient methods have been developed, see
\eg~\cite{SA89tra,Wang92tra,FM94tra,horaud1995ijrr,Strobl06RSS,ackerman2013probabilistic}. 

In this paper, we are interested, not merely in solving for $\bmX$,
but more comprehensively, in evaluating the \emph{covariance} of
$\bmX$ from those of $\bmA$ and $\bmB$, where $\bmA$ and $\bmB$ are
now \emph{randomly perturbed} transformation matrices. This
fine-grained information is critical in high-precision robotics
applications for several reasons. 

\subsection*{Motivations}

The uncertainty of the object pose estimation comes from three main
sources: (i) the uncertainty of the object pose estimation in the
camera frame, (ii) the uncertainty of the hand-eye calibration, and
(iii) the uncertainty of the robot end-effector positioning. In
practice, source (ii) arguably contributes the most: for instance, a
tiny orientation error of 0.05 degrees in the hand-eye calibration
already implies an error of 0.6 mm in object position if the latter is
70 cm away from the camera (typical viewing distance for commodity 3D
cameras). In turn, having a precise knowledge of the uncertainty of
the object pose estimation is critical:
\begin{itemize}
\item In high-precision manufacturing, it is important, not only to
  know the pose of an object, but also to \emph{guarantee} that the
  pose estimation error is within some \emph{tolerance}. For instance,
  when drilling holes in the fuselage of an aircraft, the hole
  position tolerance is 0.5 mm -- which would be violated by an error
  of 0.05 degrees in the hand-eye calibration, even when assuming that
  the object pose estimation in the camera frame is perfect (see
  above);
\item The precise knowledge of the object pose covariance
  \emph{matrix} allows one to intelligently refine the object pose
  estimation by other perception modes. For instance, in visuo-tactile
  sensor fusion~\cite{anna11tr}, knowing that the covariance of the
  object pose is comparatively large in the translation along, say,
  the X-axis will prompt us to touch the object along that axis in
  order to best reduce the uncertainty.
\end{itemize}

In addition, knowing the covariance of $\bmX$ allows improving the
calibration process itself, by \eg{} choosing the appropriate number of
measurements to achieve a desired level of precision, or choosing the
appropriate matrices $\bmA$ and $\bmB$ that minimize the covariance of
$\bmX$.

\begin{figure}[t]
  \centering
  \includegraphics[width=0.33\textwidth]{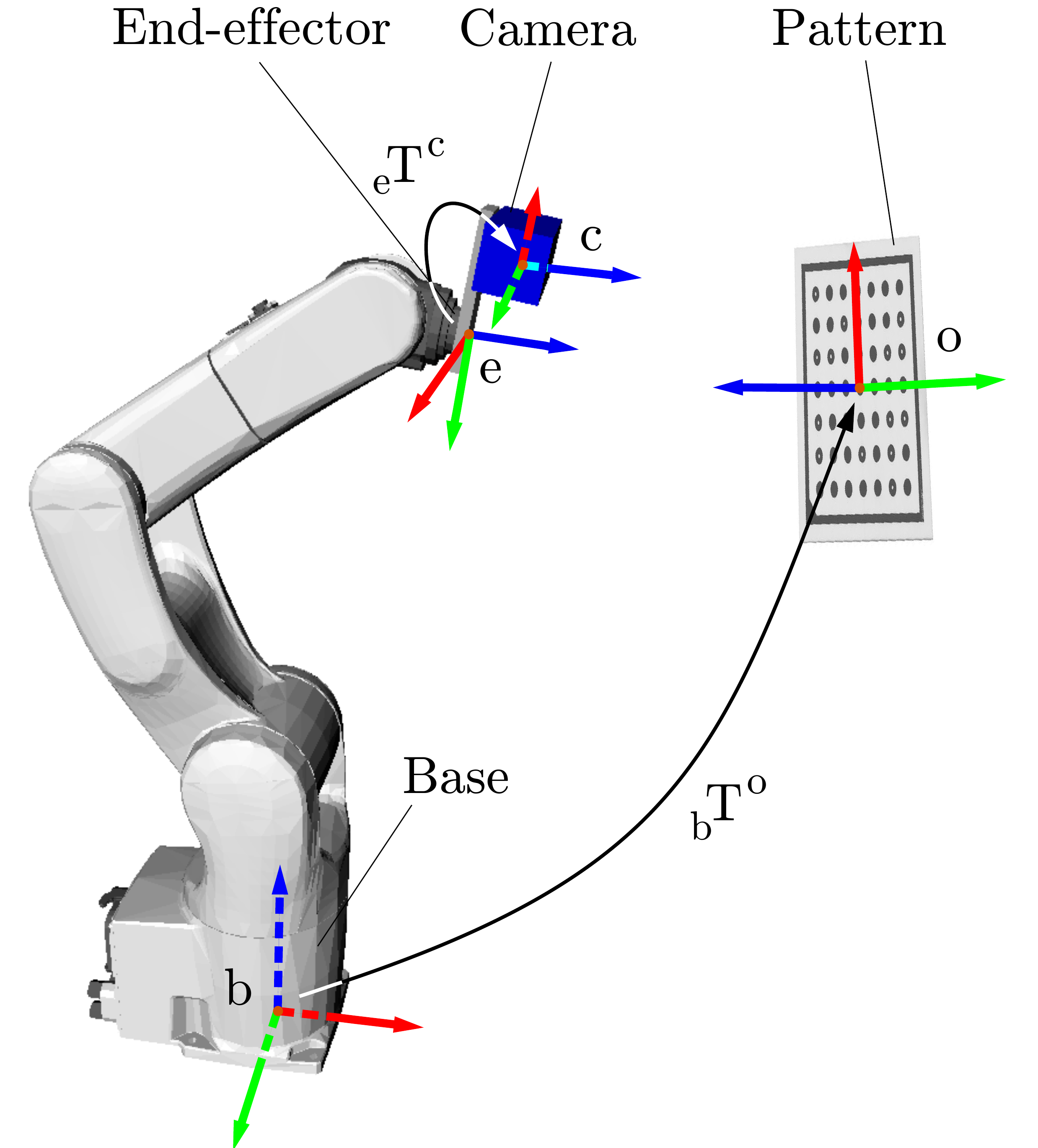}
  \caption{The hand-eye calibration problem consists in identifying
    the rigid-body transformation $\tf{\bfe}{\bmT}{\bfc}$ between a
    camera mounted on the end-effector of a robot and the end-effector
    itself.}
  \label{fig:method}
\end{figure}

\subsection*{Related works}

Finding the covariance of $\bmX$ is challenging for several
reasons. First, as $\bmX$, $\bmA$ and $\bmB$ represent rigid-body
transformations, they live in $\SE$, a subset of the space of
$4\times 4$ matrices endowed with a non-trivial Lie group
structure~\cite{murray1994book}. Second, how to represent and
calculate uncertainties in $\SE$ is by itself a complex issue, which
has prompted advanced mathematical
developments~\cite{Barfoot14tr}. Finally, merely solving for $\bmX$ in
$\bmA\bmX = \bmX\bmB$ is already a difficult
problem~\cite{FM94tra,horaud1995ijrr,Strobl06RSS},
let alone evaluating the uncertainty of the solution.

There are a number of works dealing with the uncertainty of hand-eye
calibration. In \cite{Tsai89tra}, based on a sensitivity analysis of
closed-form solutions, some critical factors and criteria influencing
the accuracy of the result are analyzed. For instance, one may try to
maximize the angle between rotation axes of relative movement to
reduce the influence on error in rotation, or to minimize the distance
between the optical center of the camera and the calibration pattern
to reduce the influence on error in translation. Based on this
analysis, Shi~\etal{} \cite{shi2005pria} present a algorithm to select
movement pairs automatically from a series of measurements to reduce
the error of the estimate. Schmidt~\etal{} also introduce similar
approach based on a vector quantization method
\cite{schmidt2008ijrr}. In \cite{aron2004ismar}, Aron~\etal{} present
an error estimation method of the rotation part of $\bmX$ based on an
Euler angles parameterization. The authors do not discuss how that
error propagates to the translation part of $\bmX$ and their vision
tracking measurements are also assumed to be noise-free. More
fundamentally, the Euler angles formulation, as opposed to the $\SE$
formulation, is well-known to involve singularities.

The idea of estimating explicitly uncertainties in the system is by no
means new. Many have studied the problem of uncertainty in the camera model
(intrinsic and extrinsic parameters)~\cite{hartley03book} and the propagation of uncertainties through
the camera model~\cite{Barfoot14tr}. However, we stress that this work
is different in that it focuses
on the hand-eye transformation and its uncertainty.

\subsection*{Contribution and organization of the paper}

It can be noted that none of the aforementioned works has provided a
derivation of the covariance of $\bmX$, which is ultimately the most
generic and relevant quantification of the uncertainty of the hand-eye
calibration process. The goal of this paper is to rigorously work out
such a derivation. Specifically, we transpose methods for forward and
backward propagation of covariance~\cite{hartley03book} into the
framework of uncertainty in $\SE$ proposed by Barfoot and
Furgale~\cite{Barfoot14tr}. The structure of the hand-eye calibration
equation raises specific technical difficulties, which we shall
address in detail.

The remainder of the paper is organized as follows. In
Section~\ref{sec:background}, we state the hand-eye calibration
problem and introduce the mathematical background of the work, which
includes the representation of uncertainty in $\SE$, and methods for
forward and backward propagation of covariance. In
Section~\ref{sec:derivation}, we present our method to estimate the
rotation and translation parts of the hand-eye transformation matrix
and their associated covariance matrices. In Section
\ref{sec:experiments}, we show that the method can indeed predict
with excellent precision these covariances in synthetic and real
calibration datasets, and uses this information to compute the
covariance of the object pose estimation in a real setting. Finally,
in Section~\ref{sec:conclusion}, we conclude by discussing the
advantages and drawbacks of our approach and sketch some future
research directions.

\section{Background}
\label{sec:background}

\subsection{Formulation of the hand-eye calibration problem}
\label{sec:hand-eye}

The classical hand-eye calibration method consists in looking at a
fixed pattern from two different viewpoints, say 1 and~2, giving rise
to the following equation

\begin{equation}
  \label{eq:ax=xb2}
  \tf{\bfb}{\bmT}{\bfe}_{1}\tf{\bfe}{\bmT}{\bfc}\tf{\bfc}{\bmT}{\bfo}_{1}
  = \tf{\bfb}{\bmT}{\bfo} =
  \tf{\bfb}{\bmT}{\bfe}_{2}\tf{\bfe}{\bmT}{\bfc}\tf{\bfc}{\bmT}{\bfo}_{2},
\end{equation}
where
\begin{itemize}
\item $\tf{\bfb}{\bmT}{\bfe}_{i}$ is the transformation of the
  end-effector with respect to the fixed robot base at configuration
  $i$;
\item $\tf{\bfe}{\bmT}{\bfc}$ is the constant transformation of the
camera with respect to the end-effector;
\item $\tf{\bfc}{\bmT}{\bfo}_{i}$ is the transformation of the pattern
  (object) with respect to the camera at configuration $i$;
\item $\tf{\bfb}{\bmT}{\bfo}$ is the constant transformation of the
  pattern with respect to the robot base (see Fig.~\ref{fig:method}).
\end{itemize}

Next, one can transform the above equation into

\begin{equation}
  \label{eq:ax=xb1}
  {\tf{\bfb}{\bmT}{\bfe}_{2}}^{-1} \tf{\bfb}{\bmT}{\bfe}_{1} \tf{\bfe}{\bmT}{\bfc}
  =
  \tf{\bfe}{\bmT}{\bfc} \tf{\bfc}{\bmT}{\bfo}_{2} {\tf{\bfc}{\bmT}{\bfo}_{1}}^{-1},
\end{equation}
which has the form of $\bmA\bmX = \bmX\bmB$, where
$\bmX:=\tf{\bfe}{\bmT}{\bfc}$ is the unknown hand-eye transformation,
and $\bmA:={\tf{\bfb}{\bmT}{\bfe}_{2}}^{-1} \tf{\bfb}{\bmT}{\bfe}_{1}$
and $\bmB:=\tf{\bfc}{\bmT}{\bfo}_{2} {\tf{\bfc}{\bmT}{\bfo}_{1}}^{-1}$
can be computed from respectively the robot kinematics and pattern pose
estimation~\cite{FM94tra}. Next, if the fixed pattern is viewed from a
large number of viewpoints, one can collect many different $\bmA$’s
and $\bmB$’s. Suppose that we have a set of $k$ measurements
${(\bmA_{1},\bmB_{1}),(\bmA_{2},\bmB_{2}),...,(\bmA_{k},\bmB_{k})}$. Since
in practice these measurements are perturbed by actuator/sensor noise,
the exact solution for the set of $k$ equations
$\bmA_{i}\bmX = \bmX\bmB_{i}$ will not exist. Instead, the problem is
commonly framed as an optimization problem in which $\bmX$ is found as
the transformation that ``best'' fits the $k$ equalities.

Note that sometimes the camera may not be mounted on the
end-effector but on a fixed stand. In this case, finding the relative transformation between
the camera and the robot base can also be formulated
as the $\bmA\bmX=\bmX\bmB$ problem and can be treated by the same method.

\subsection{Representation of rigid-body transformations and of their
  uncertainties}
\label{subsec:poseandnoiserepresentation}

We choose to represent rigid-body transformations as elements of the
Special Euclidean group $\SE$~\cite{murray1994book}. To model the
uncertainty on $\SE$, we adopt the framework proposed
in~\cite{Barfoot14tr}. As there is in general no bi-invariant distance
on $\SE$~\cite{PR97acm}, solving for the rotation and translation
components of $\bmX$ simultaneously would in any case require an
arbitrary rotation/translation weighting. Instead, we choose to solve
them separately, which entails a number of
simplifications~\cite{FM94tra}. As a consequence, the uncertainties of
the rotation and the translation parts are also modeled separately.

Specifically, we assume that the rotation parts of the observations
$\bmA_i$ and $\bmB_i$ are corrupted by Gaussian noise as follows
\begin{eqnarray}
\label{eq:rotationnoisemodelA}
\bmR_{\bmA_{i}} = \exp([\bfxi_{\bmR_{\bmA}i}])\bar{\bmR}_{\bmA_{i}},\\
\label{eq:rotationnoisemodelB}
\bmR_{\bmB_{i}} = \exp([\bfxi_{\bmR_{\bmB} i}])\bar{\bmR}_{\bmB_{i}},
\end{eqnarray}
where $\bar{\bmR}_{\bmA_{i}},\bar{\bmR}_{\bmB_{i}}\in \SO$ are the
means of $\bmR_{\bmA_{i}},\bmR_{\bmB_{i}}$, and
$\bfxi_{\bmR_{\bmA} i},\bfxi_{\bmR_{\bmB} i}\in\bbR^3$ are zero-mean
Gaussian perturbations with covariance matrices
$\bfSigma_{\bmR_{\bmA} i},\bfSigma_{\bmR_{\bmB} i}$,
respectively.

The translation parts of the $\bmA_i$ and $\bmB_i$ are corrupted as
follows 
\begin{eqnarray}
\label{eq:translationnoisemodelA}
\bft_{\bmA_{i}} = \bfxi_{\bft_{\bmA} i} + \bar{\bft}_{\bmA_{i}},\\
\label{eq:translationnoisemodelB}
\bft_{\bmB_{i}} = \bfxi_{\bft_{\bmB} i} + \bar{\bft}_{\bmB_{i}},
\end{eqnarray}
where $\bar{\bft}_{\bmA_{i}},\bar{\bft}_{\bmB_{i}}\in\bbR^3$ are the
means of $\bft_{\bmA_{i}},\bft_{\bmB_{i}}$, and
$\bfxi_{\bft_{\bmA} i},\bfxi_{\bft_{\bmB} i} \in\bbR^3$ are zero-mean
Gaussian perturbations with covariance matrices
$\bfSigma_{\bft_{\bmA} i},\bfSigma_{\bft_{\bmB} i}$, respectively.

Note that the above assumptions imply that rotation and translation
noises are independent.

\subsection{Forward and backward propagation of covariance}
\label{sec:fwdbwdpropagation}

\emph{Forward propagation}. Let $\bfP$ be a random vector in $\bbR^M$
with mean $\bar{\bfP}$ and covariance matrix $\bfSigma$. Consider a
function $\bff: \bbR^M \rightarrow \bbR^N$ that is differentiable in a
neighbourhood of $\bar\bfP$. Then, at the first order of
approximation, $\bff(\bfP)$ is a random variable with mean
$\bff(\bar\bfP)$ and covariance matrix
\begin{equation}
\label{eq:fwdpropagation}
\bfSigma_{\bff}=\bfJac\bfSigma\bfJac^\top,
\end{equation}
where $\bfJac$ is the Jacobian matrix of
$\bff$ at $\bar\bfP$.

\emph{Backward propagation}. Assume now that $\bfP$ (the parameter) is
unknown, but that $\bfV:=\bff(\bfP)$ (the measurement) is known and
determined to be a random variable with mean $\bar{\bfV}$ and
covariance matrix $\bfSigma_{\bfV}$. Then the best estimate for $\bfP$
is given by
\[
\bfP^* := \min_{\bfP} \|\bfV-\bff(\bfP)\|_{\bfSigma_{\bfV}}.
\]
To estimate the covariance of $\bfP$, one can approximate $\bff$ by an
affine function
$\bff(\bfP) = \bff(\bar\bfP) + \bfJac(\bfP - \bar{\bfP})$, which yields
{\small\begin{equation}
  \|\bfV - \bff(\bfP)\|_{\bfSigma_{\bfV}} = 
  \|(\bfV - \bar{\bfV})-\bfJac(\bfP-\bar{\bfP})\|_{\bfSigma_{\bfV}}. 
\end{equation}}
Using the weighted pseudo-inverse, one has
{\small
\begin{equation}
  \bfP^* - \bar{\bfP} =
  (\bfJac^\top{\bfSigma_{\bfV}}^{-1}\bfJac)^{-1}
  \bfJac^\top{\bfSigma_{\bfV}}^{-1}(\bfV-\bar{\bfV}).
\end{equation}}
From~(\ref{eq:fwdpropagation}), the covariance of $\bfP$ can now be
approximated at the first order by 
{\small
  \begin{eqnarray}
    \label{eq:bwdpropagation}
    \bfSigma^*&=&[(\bfJac^\top{\bfSigma_{\bfV}}^{-1}\bfJac)^{-1}\bfJac^\top{\bfSigma_{\bfV}}^{-1}]
                    {\bfSigma_{\bfV}}\nonumber\\
                &&[(\bfJac^\top{\bfSigma_{\bfV}}^{-1}\bfJac)^{-1}\bfJac^\top{\bfSigma_{\bfV}}^{-1}]^\top 
                    \nonumber\\
                &=& (\bfJac^\top{\bfSigma_{\bfV}}^{-1}\bfJac)^{-1}.
  \end{eqnarray}}
In practice, when performing an iterative least-squares optimization,
one can use~(\ref{eq:bwdpropagation}) at the last iteration to obtain
the estimation of the covariance of $\bfP$.

Note that the quality of the approximations given by
Equations~(\ref{eq:fwdpropagation}) and~(\ref{eq:bwdpropagation})
depends in particular on the quality of the linear approximation of
$\bff$.

\section{Derivation of the covariance of \texorpdfstring{$\bmX$}{X}}
\label{sec:derivation}

Equation $\bmA_{i}\bmX = \bmX\bmB_{i}$ can be decomposed as
\begin{gather}
\label{eqn:handeyerotation}
\bmR_{\bmA i}\bmR= \bmR\bmR_{\bmB i},\\
\label{eqn:handeyetranslation}
\bmR_{\bmA i}\bft + \bft_{\bmA i} = \bmR\bft_{\bmB i}+\bft,
\end{gather} 
where $\bmR,\bft$ denote respectively the rotation and translation
parts of $\bmX$.

\subsection{Covariance of the rotation part of \texorpdfstring{$\bmX$}{X}}

We first consider the rotation part $\bmR$ of $\bmX$. Let
$[\bfalpha_{i}],[\bfbeta_{i}]\in\so$ denote the logarithms of
$\bmR_{\bmA i}$ and $\bmR_{\bmB i}$ respectively, \ie
\begin{gather}
  \label{eq:logarithm} [\bfalpha_{i}] := \log{\bmR_{\bmA i}}, \quad
  [\bfbeta_{i}] := \log{\bmR_{\bmB i}}.
\end{gather}
Note that the covariance matrices of $\bfalpha_{i}$ and $\bfbeta_{i}$
can be obtained by applying the forward propagation of covariance
\begin{gather}
  \bfSigma_{\bfalpha_{i}} =
  \bmJ(\bfalpha_{i})^{-1}\bfSigma_{\bmR_{\bmA} i}\bmJ(\bfalpha_{i})^{-1\top},\\
  \bfSigma_{\bfbeta_{i}} =
  \bmJ(\bfbeta_{i})^{-1}\bfSigma_{\bmR_{\bmB} i}\bmJ(\bfbeta_{i})^{-1\top},
\end{gather}
where $\bmJ(\bfalpha_{i})$ denotes the (left) Jacobian of $\SO$ at
$\bfalpha_i$, see \cite{Barfoot14tr} for more details.

Next, via logarithm mapping, equation (\ref{eqn:handeyerotation}) can
be written as
\begin{equation}
\log{\bmR_{\bmA i}} =\log{\bmR\bmR_{\bmB i} \bmR^\top}=\bmR[\bfbeta_{i}]\bmR^\top.
\end{equation}
Applying the rule $\bmR[\bfomega]\bmR^\top=[\bmR\bfomega]$ for
$\bmR\in\SO$ and $[\bfomega]\in\so$, one has
\begin{equation}
\label{eq:rotationinexpform}
\bfalpha_{i}=\bmR\bfbeta_{i}.
\end{equation}
\label{sec:uncertaintyofsolution}
In order to use the uncertainty model in $\SO$, we define a random
variable $\bfxi_{\bmR}$ that represents the difference between $\bmR$
and the current estimate $\hat\bmR$ by
\[
\bmR = \exp([\bfxi_{\bmR}])\hat\bmR.
\]
Next, to apply the backward propagation of covariance, one needs the
measurement vectors $\bfalpha_i$ and $\bfbeta_i$ to appear on the same
side of the equation. To achieve this without making it too complex,
we use a trick from~\cite{hartley03book}, which consists in
``copying'' the $\bfbeta_i$'s on both sides, as follows
\begin{equation}
\underbrace{\left(
  \begin{array}{c}
    \bfbeta_1\\
    \bfalpha_1\\
    \vdots\\
    \bfbeta_k\\
    \bfalpha_k\\
  \end{array}
\right)}_{\bfV} =
\underbrace{
\left(
  \begin{array}{c}
    \bfbeta_1\\
    \exp([\bfxi_{\bmR}])\hat{\bmR}\bfbeta_1\\
    \vdots\\
    \bfbeta_k\\
    \exp([\bfxi_{\bmR}])\hat{\bmR}\bfbeta_k\\
  \end{array}
\right)}_{\bff(\bfP)}.
\end{equation}
Now, the measurement vector is given by $\bfV:=(\bfV_1,\dots,\bfV_k)$,
where $\bfV_i:=(\bfbeta_i,\bfalpha_i)$, and the parameter vector is
given by $\bfP:=(\bfxi_{\bmR},\bfbeta_1,\dots,\bfbeta_k)$.

Since the noise of $\bfalpha_{i}$'s and $\bfbeta_{i}$'s are
independent ($\bfalpha$ is caused by robot kinematics while
$\bfbeta_i$ is caused by object pose estimation in the camera frame),
the covariance matrix of the measurement vector is given by
\begin{gather}
  \bfSigma_{\bfV}:=\diag(\bfSigma_{\bfV_{1}},\bfSigma_{\bfV_{2}},...,\bfSigma_{\bfV_{k}}),
\end{gather} 
with $\bfSigma_{\bfV_{i}}:=\diag(\bfSigma_{\bfbeta_{i}},\bfSigma_{\bfalpha_{i}})$.

Now, the covariance-weighted minimization is given by
\[
\min_{\bfP} \|\bfV - \bff(\bfP) \| = 
\min_{\bfP} \sum_{i}^{k} (\bfV_i-\bff(\bfP)_i)^\top
{\bfSigma_{\bfV_i}^{-1}}(\bfV_i-\bff(\bfP)_i).
\]
This minimization problem can be solved by iteratively updating the
estimate of the parameter vector by the rules
\begin{eqnarray}
\hat{\bmR}_{(j+1)} &=& 
\exp([\bfxi_{\bmR}]) \hat{\bmR}_{(j)},\\
\hat{\bfbeta}_{i (j+1)} &=& \hat{\bfbeta}_{i (j)}+\bfdelta_{\bfbeta_{i}},
\end{eqnarray}
where at each step $(j)$ the update vector
$\bfdelta := (\bfxi_{\bmR},\bfdelta_{\bfbeta_{1}},\dots,\bfdelta_{\bfbeta_{k}})$
is found by solving the normal equation
\begin{equation}
\label{eq:normalequationrot}
\bfJac_{\bff}^\top {\bfSigma_{\bfV}^{-1}}\bfJac_{\bff}\bfdelta 
= \bfJac_{\bff}^\top {\bfSigma_{\bfV}^{-1}}(\bfV-\bff(\bfP)).
\end{equation}
The Jacobian of $\bff$ has the form
\begin{eqnarray}
\bfJac_{\bff}&=&\left[\bfJac^{\bfxi_{\bmR}}|\bfJac^{\bfbeta}\right]\nonumber\\
\label{eqn:Jac}                 &=&\left[\begin{array}{cccccc}
                            \bfJac^{\bfxi_{\bmR}}_{1}&\vline&\bfJac^{\bfbeta}_{1}& & & \\
                            \bfJac^{\bfxi_{\bmR}}_{2}&\vline& &\bfJac^{\bfbeta}_{2}&  & \\
                            \vdots &\vline& & &\ddots&  \\
                            \bfJac^{\bfxi_{\bmR}}_{k}&\vline& & & &\bfJac^{\bfbeta}_{k}
                          \end{array}
                                                                   \right],
\end{eqnarray}
\begin{equation}
 \text{where } \bfJac^{\bfxi_{\bmR}}_{i}:=
  \left[ \begin{array}{c}
           \bf0\\-[\hat{\bmR}{\hat{\bfbeta}}_{i}]
         \end{array}
       \right], 
  \bfJac^{\bfbeta}_{i}:=
  \left[ \begin{array}{c}
           \bbI\\\hat{\bmR}
         \end{array}
       \right].
\end{equation}

The set of equations (\ref{eq:normalequationrot}) may now written in block
form as
\begin{gather}
\left[
  \begin{array}{cc} \bfJac^{\bfxi_{\bmR}\top}\bfSigma_{\bfV}^{-1}\bfJac^{\bfxi_{\bmR}}&\bfJac^{\bfxi_{\bmR}\top}\bfSigma_{\bfV}^{-1}\bfJac^{\bfbeta}\\
\bfJac^{\bfbeta\top}\bfSigma_{\bfV}^{-1}\bfJac^{\bfxi_{\bmR}}&\bfJac^{\bfbeta\top}\bfSigma_{\bfV}^{-1}\bfJac^{\bfbeta}
  \end{array}
\right]
\left(
  \begin{array}{c}
    \bfxi_{\bmR}\\\bfdelta_{\bfbeta}
  \end{array}
\nonumber
\right)\\
\label{eq:normaleqblockform}
=\left(
\begin{array}{c}
  \bfJac^{\bfxi_{\bmR}\top}\bfSigma_{\bfV}^{-1}(\bfV-\bff(\bfP))\\
  \bfJac^{\bfbeta\top}\bfSigma_{\bfV}^{-1}(\bfV-\bff(\bfP))
\end{array}
\right).
\end{gather}

To simplify the left-hand side of~(\ref{eq:normaleqblockform}), let
\begin{eqnarray}
\label{eq:simplifyUforrotation}
\bmU:=&\bfJac^{\bfxi_{\bmR}\top}\bfSigma_{\bfV}^{-1}\bfJac^{\bfxi_{\bmR}}=\sum_{i}^{k}\bfJac^{\bfxi_{\bmR}\top}_{i}\bfSigma_{\bfV_{i}}^{-1}\bfJac_{i}^{\bfxi_{\bmR}},\\
\label{eq:simplifyWforrotation}
\bmW:=&\bfJac^{\bfxi_{\bmR}\top}\bfSigma_{\bfV}^{-1}\bfJac^{\bfbeta}=\left[\bmW_{1},...,\bmW_{k}\right],\\
\label{eq:simplifyZforrotation}
\bmZ:=&\bfJac^{\bfbeta\top}\bfSigma_{\bfV}^{-1}\bfJac^{\bfbeta}
=\diag(\bmZ_{1},...,\bmZ_{k}),\\
\text{ where }  
\bmW_{i}:=&\bfJac^{\bfxi_{\bmR}\top}_{i}\bfSigma_{\bfV_{i}}^{-1}\bfJac_{i}^{\bfbeta}
\text{ and }
\bmZ_{i}:=\bfJac^{\bfbeta\top}_{i}\bfSigma_{\bfV_i}^{-1}\bfJac_i^{\bfbeta}.
\end{eqnarray}

As for the right-hand side of~(\ref{eq:normaleqblockform}), let  
\begin{eqnarray}
\bfepsilon_{\bfxi_{\bmR}}
:=&\bfJac^{\bfxi_{\bmR}\top}\bfSigma_{\bfV}^{-1}(\bfV-\bff(\bfP))\nonumber\\
=&\sum_{i}^{k}\bfJac^{\bfxi_{\bmR}\top}_{i}\bfSigma_{\bfV_i}^{-1}(\bfV_i-\bff(\bfP)_i),\\
\bfepsilon_{\bfbeta} 
:=& \bfJac^{\bfbeta\top}\bfSigma_{\bfV}^{-1}(\bfV-\bff(\bfP)) 
=
({\bfepsilon_{\bfbeta}}_{1},\dots,{\bfepsilon_{\bfbeta}}_{k}),\\ \text{
  where }\bfepsilon_{\bfbeta_i}:=&\bfJac^{\bfbeta\top}_i\bfSigma_{\bfV_i}^{-1}(\bfV_i-\bff(\bfP)_i).
\end{eqnarray}

To solve equations~(\ref{eq:normaleqblockform}), one can left-multiply
both sides by {\footnotesize$\left[
  \begin{array}{cc}
    \bbI&\bmW\bmZ^{-1}\\
    \bf0&\bbI
  \end{array}
\right]$}, which yields
\begin{eqnarray}
\label{eq:normaleqR}
(\bmU-\bmW\bmZ^{-1}\bmW^\top)\bfxi_{\bmR}&=&\bfepsilon_{\bfxi_{\bmR}}-\bmW\bmZ^{-1}\bfepsilon_{\bfbeta},\\
\bmZ\bfdelta_{\bfbeta}&=&\bfepsilon_{\bfbeta}-\bmW^\top\bfxi_{\bmR}.
\end{eqnarray}
The above equations can now be solved to find the updating vectors
$\bfxi_{\bmR}$ and $\bfdelta_{\bfbeta}$.

Applying backward propagation of covariance, a first-order
approximation of the covariance of $\bfP$ is given the following
matrix, taken at the last iteration,
\begin{eqnarray}
\bfSigma^*
&=&(\bfJac_{\bff}^\top{\bfSigma_{\bfV}}^{-1}\bfJac_{\bff})^{-1}\nonumber\\
&=&
\left[
  \begin{array}{cc}
    \bfSigma_{\bmR}&-\bfSigma_{\bmR}\bmW\bmZ^{-1}\\
    -(\bfSigma_{\bmR}\bmW\bmZ^{-1})^\top&(\bmW\bmZ^{-1})^\top \bfSigma_{\bmR}\bmW\bmZ^{-1}+\bmZ^{-1}
  \end{array}
\right].\nonumber
\end{eqnarray}
The covariance of $\bfxi_{\bmR}$ is given by the top-left block of
$\bfSigma^*$, that is: 
\begin{equation}
\label{eq:sigmarotation}
\bfSigma_{\bmR}= (\bmU-\bmW\bmZ^{-1}\bmW^\top)^{-1}=(\bmU-\sum_{i}^{k}\bmW_{i}{\bmZ_{i}}^{-1}{\bmW_{i}}^\top)^{-1}.
\end{equation}

\subsection{Covariance of the translation part of \texorpdfstring{$\bmX$}{X}}
We now consider the translation part $\bft$ of $\bmX$. Let
$\bfq_{i}:=\bmR\bft_{\bmB_{i}}-\bft_{\bmA_{i}}$. Equations~(\ref{eqn:handeyetranslation}) 
can be written as
\begin{equation}
\bfq_{i}=(\bmR_{\bmA_{i}}-\bbI)\bft.
\end{equation}

Note that the covariance matrices of $\bfq_{i}$ can be
approximated by applying the forward propagation of covariance
\begin{equation}
\bfSigma_{\bfq_{i}}=\bfSigma_{\bft_{\bmA}}+{\bmR^*}\bfSigma_{\bft_{\bmB}}{\bmR^*}^\top+[{\bmR^*}{\bft_{\bmB}}_{i}]\bfSigma_{\bmR}[{\bmR^*}{\bft_{\bmB}}_{i}]^\top,
\label{eq:covarianceofq}
\end{equation}

where ${\bmR^*}$ is the optimal rotation found in the previous
section, and $\bfSigma_{\bmR}$ is the corresponding covariance.

Applying the same trick as previously, we ``copy'' the $\bmR_{\bmA i}$'s
on both sides of the equation, as follows
\begin{equation}
\underbrace{\left(
  \begin{array}{c}
    \bmR_{\bmA_1}\\
    \bfq_1\\
    \vdots\\
    \bmR_{\bmA_k}\\
    \bfq_k\\
  \end{array}
\right)}_{\bfV} =
\underbrace{
\left(
  \begin{array}{c}
    \bmR_{\bmA_1}\\
    (\bmR_{\bmA_1}-\bbI)\hat{\bft}\\
    \vdots\\
    \bmR_{\bmA_k}\\
    (\bmR_{\bmA_k}-\bbI)\hat{\bft}\\
  \end{array}
\right)}_{\bff(\bfP)}.
\end{equation}

Now the measurement vector is given by $\bfV:=(\bfV_1,\dots,\bfV_k)$,
where $\bfV_i:=(\bmR_{\bmA_i},\bfq_{i})$, and the parameter vector is
given by $\bfP:=(\bft,\bmR_{\bmA_1},\dots,\bmR_{\bmA_k})$.

Since computing the cross-variance of $\bmR_{\bmA_i}$ and $\bfq_{i}$
would be too complex, we simply assume them to be independent. The
covariance matrix of the measurement vector is then given by
\begin{equation}
{\bfSigma_{\bfV}}:=\diag(\bfSigma_{\bfV_{1}},
\bfSigma_{\bfV_{2}},\dots,\bfSigma_{\bfV_{k}}),
\end{equation} where
${\bfSigma_{\bfV_{i}}}:=\diag(\bfSigma_{{\bmR_{\bmA}}_{i}},
\bfSigma_{\bfq_{i}})$. 

Now, the covariance-weighted minimization is given by
\[\min_{\bfP} \|\bfV - \bff(\bfP) \| = 
\min_{\bfP} \sum_{i}^{k} (\bfV_i-\bff(\bfP)_i)^\top
{\bfSigma_{\bfV_i}^{-1}}(\bfV_i-\bff(\bfP)_i).\]
We solve this by iteratively updating the estimate of the parameter
vector by the rules
\begin{eqnarray}
\hat{\bft}_{(j+1)}&=&\hat{\bft}_{(j)}+\bfdelta_{\bft},\\
\hat{\bmR}_{\bmA_{i}(j+1)}&=&\exp([{\bfxi_{\bmR_{\bmA}}}_{i}])\hat{\bmR}_{\bmA_{i}(j)},
\end{eqnarray}
where at each step $(j)$ the update vector $\bfdelta
:=(\bfdelta_{\bft},{\bfxi_{\bmR_{\bmA}}})=
(\bfdelta_{\bft},\bfxi_{\bmR_{\bmA_1}},\dots,\bfxi_{\bmR_{\bmA_k}})$
is found by solving the normal equation
\begin{equation}
\label{eq:normalequationtrans}
\bfJac_{\bff}^\top \bfSigma_{\bfV}^{-1}\bfJac_{\bff}\bfdelta = \bfJac_{\bff}^\top \bfSigma_{\bfV}^{-1}(\bfV-\bff(\bfP)).
\end{equation}
 The Jacobian matrix has the form
 \begin{eqnarray}
\bfJac_{\bff}=&\left[\bfJac^{\bft}|\bfJac^{{\bfxi_{\bmR_{\bmA}}}}\right],
\\\text{where }
\bfJac^{\bft}:=&(\bfJac^{\bft}_{1},\dots,\bfJac^{\bft}_{k}),\\
\bfJac^{{\bfxi_{\bmR_{\bmA}}}}:=&\diag(\bfJac^{{\bfxi_{\bmR_{\bmA}}}}_{1},...,\bfJac^{{\bfxi_{\bmR_{\bmA}}}}_{k}),\\
\bfJac^{\bft}_{i}:=&\left[\begin{array}{c}
                            \bf0\\\hat{\bmR}_{\bmA_{i}}-\bbI
                          \end{array}
  \right],\\
\bfJac^{{\bfxi_{\bmR_{\bmA}}}}_{i}:=&
  \left[ \begin{array}{c}
           \bbI\\-[\hat{\bmR}_{\bmA_{i}}\hat{\bft}_{\bmX}]
         \end{array}
  \right].
\end{eqnarray}

For the rest of the derivation, we following the same procedure as
previously derived. One thus can obtain the update vectors from
\begin{eqnarray}
(\bmU-\bmW\bmZ^{-1}\bmW^\top)\bfdelta_{\bft}=&\bfepsilon_{\bft}-\bmW\bmZ^{-1}\bfepsilon_{\bfxi_{\bmR_{\bmA}}},\\
\bmZ{\bfxi_{\bmR_{\bmA}}}=&\bfepsilon_{\bfxi_{\bmR_{\bmA}}}-\bmW^\top\bfdelta_{\bft},
\end{eqnarray}
\begin{eqnarray}
\label{eq:simplifyUfortrans}
\text{where }
  \bmU:=&\sum_{i}^{k}\bfJac^{\bft\top}_{i}\bfSigma_{\bfV_{i}}^{-1}\bfJac_{i}^{\bft},\\
\label{eq:simplifyWfortrans}
\bmW:=&\left[\bmW_{1},\dots,\bmW_{k}\right]  \text{ with }
\bmW_{i}:=\bfJac^{\bft\top}_{i}\bfSigma_{\bfV_{i}}^{-1}\bfJac_{i}^{{\bfxi_{\bmR_{\bmA}}}},\\
\label{eq:simplifyZfortrans}
\bmZ:=&\mathrm{diag}(\bmZ_{1},\dots,\bmZ_{k})  \text{ with }
\bmZ_{i}:=\bfJac^{{\bfxi_{\bmR_{\bmA}}}\top}_{i}\bfSigma_{\bfV_{i}}^{-1}\bfJac_{i}^{{\bfxi_{\bmR_{\bmA}}}},\\
\bfepsilon_{\bft} :=&
\sum_{i}^{k}\bfJac^{\bft\top}_{i}\bfSigma_{\bfV_{i}}^{-1}(\bfV_i-\bff(\bfP)_i),\\
\bfepsilon_{\bfxi_{\bmR_{\bmA}}} :=&
(\bfepsilon_{\bfxi_{\bmR_{\bmA_1}}},\dots,\bfepsilon_{\bfxi_{\bmR_{\bmA_k}}}),\\
\text{ with } \bfepsilon_{\bfxi_{\bmR_{\bmA_i}}}=&\bfJac^{{\bfxi_{\bmR_{\bmA}}}\top}_{i}\bfSigma_{\bfV_{i}}^{-1}(\bfV_i-\bff(\bfP)_i).
\end{eqnarray}
At the last iteration, a first-order approximation of the covariance
matrix of $\bft$ is given by
{\small
\begin{equation}
\label{eq:sigmatranslation}
\bfSigma_{\bft}=(\bmU-\bmW\bmZ^{-1}\bmW^\top)^{-1}=(\bmU-\sum_{i}^{k}\bmW_{i}{\bmZ_{i}}^{-1}{\bmW_{i}}^\top)^{-1}.
\end{equation}}
\section{Experiments}
\label{sec:experiments}

We now validate the proposed method by comparing the covariance
predicted by the method and that obtained from Monte-Carlo
simulations, using synthetic and real calibration data. Using the
covariance of $\bmX$, we are then in a position to compute the
covariance of the object pose estimation in a real setting. Our
implementation is open-source and is available at
\url{https://github.com/dinhhuy2109/python-cope}.

\subsection{Synthetic calibration data}

To generate synthetic data, we start by selecting a random
transformation matrix $\bar\bmX = (\bar{\bmR}, \bar{\bft})$, which
serves as the true hand-eye transformation. We then generate $M=1000$
dataset, each dataset comprising $k=30$ corrupted pairs
$(\bmA_i,\bmB_i)_{i\in [1,k]}$. Each corrupted pair is generated as
follows. First, we generate a random \emph{uncorrupted} pair
$(\bar\bmA_i,\bar\bmB_i)$, which verifies $\bar\bmA_{i}\bar\bmX =
\bar\bmX\bar\bmB_{i}$ exactly. Next, we add noise to
$\bar\bmA_i$ and $\bar\bmB_i$ as explained in
Section~\ref{subsec:poseandnoiserepresentation}. The covariance
matrices of the noise are chosen arbitrarily as
{\small
\begin{eqnarray}
  \bfSigma_{\bmR_{\bmA_i}} &:=&\lambda \text{~diag} (0.5,0.2,0.3),\\
  \bfSigma_{\bmR_{\bmB_i}} &:=& \lambda \text{~diag}(0.7,0.2,0.8),\\
  \bfSigma_{\bft_{\bmA_i}} &:=& \lambda \text{~diag}(0.1,0.2,0.5),\\
  \bfSigma_{\bft_{\bmB_i}} &:=& \lambda \text{~diag}(0.7,0.8,0.1),
\end{eqnarray}}
where $\lambda \in \bbR$ is a scaling parameter that allows
us to change the magnitude of the uncertainties. 

At each noise level $\lambda$, we evaluate the covariance of $\bmX$
following two methods

\begin{figure}[t]
  \centering
  \includegraphics[width=0.4\textwidth]{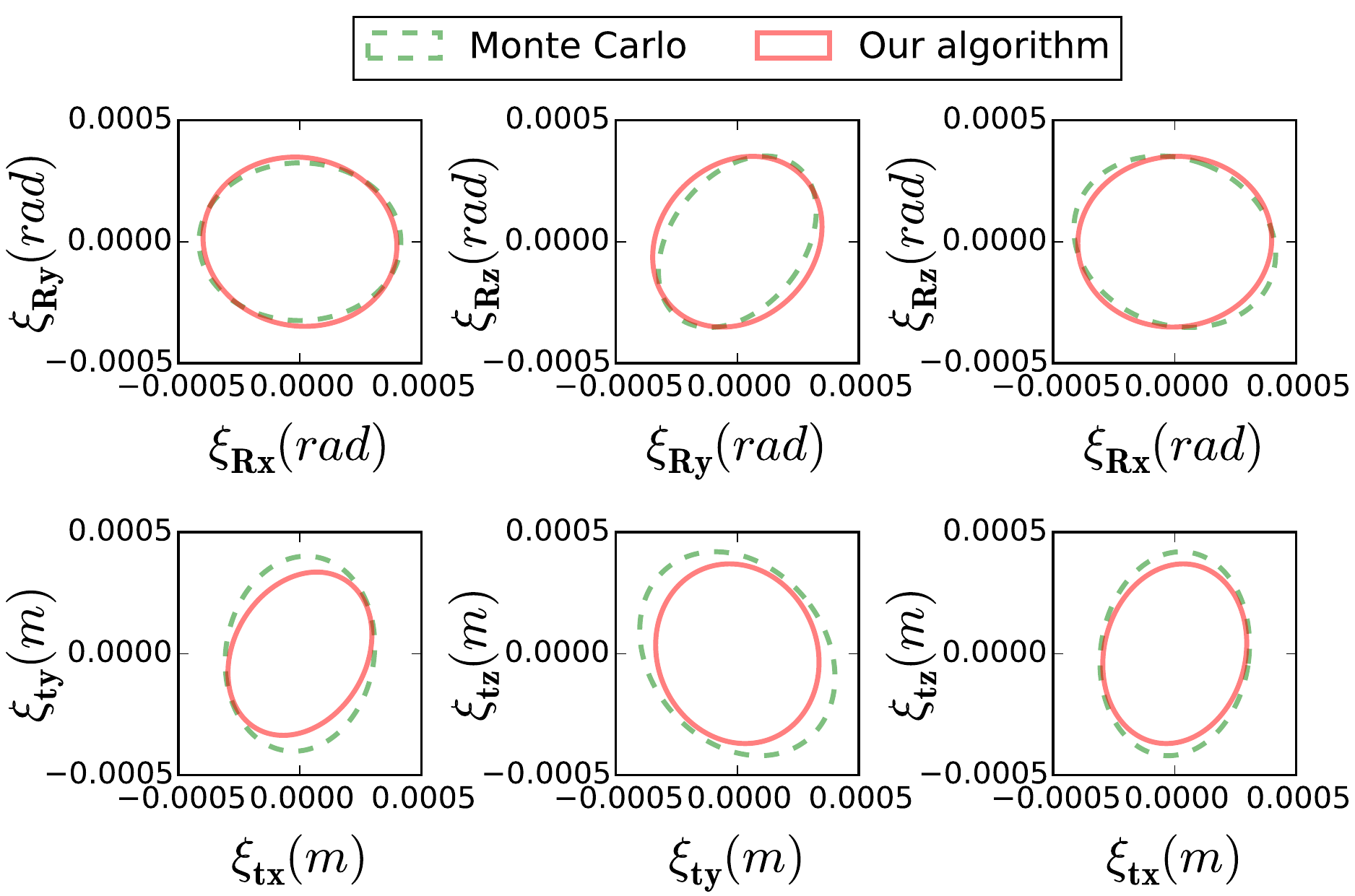}
  \caption{Projection of the one-standard-deviation covariance
    ellipsoids on pairs of axes for $\bfSigma_{\bmR}$ and
    $\bfSigma_{\bft}$, shown for Monte-Carlo and our method on
    synthetic data when $\lambda = 10^{-5}$, where $\bfxi_{\bmR x},\bfxi_{\bmR y},\bfxi_{\bmR z}$ are
    errors in rotation around X, Y, Z-axis and
    $\bfxi_{\bft x},\bfxi_{\bft y},\bfxi_{\bft z}$ are errors in
    translation along X, Y, Z-axis.}
  \label{fig:compareellipsesyn}
\end{figure}

\begin{itemize}
\item \emph{Our method:} For some dataset $m\in[1,M]$, we compute the
  covariance matrices $\bfSigma_{\bmR_m},\bfSigma_{\bft_m}$ using the
  proposed method (\underline{PR}edicted). In fact, the
  $\bfSigma_{\bmR_m},\bfSigma_{\bft_m}$ are nearly identical across
  the $M$ datasets, so the particular value of $m$ did not matter; 
\item \emph{Monte-Carlo:} For each dataset $m\in[1,M]$, we find the
  rotation and translation $\hat{\bmR}_{m}$, $\hat{\bft}_{ m}$ that
  optimally fit the $k$ hand-eye equations following the proposed method.
  We then compute the
  covariance matrices $\bfSigma_{\bmR}^\mathrm{MC}$,
  $\bfSigma_{\bft}^\mathrm{MC}$ by the
  \underline{M}onte-\underline{C}arlo method across the $M$ datasets
  as
  \begin{gather}
\label{eq:mc_rot}
  \bfSigma_{\bmR}^\mathrm{MC}:=\frac{1}{M}\sum_{m}^{M}{\bfxi_{\bmR_m}}
  \bfxi_{\bmR_m}^\top,\\
\label{eq:mc_trans}
  \bfSigma_{\bft}^\mathrm{MC}:=\frac{1}{M}\sum_{m}^{M}\bfxi_{\bft_m}
          {\bfxi_{\bft_m}}^\top,
    \end{gather}
 where $\bfxi_{\bmR_m}:=(\log(\hat{\bmR}_m \bar{\bmR}^{-1}))^{\vee}$\footnote{The $[~]$ operator
  turns $\bfxi\in\bbR^3$ into a member of the Lie algebra $\se$ (see
  equation~(\ref{eq:logarithm})), we use $\vee$ as the inverse
  operation of $[~]$.}, $\bfxi_{\bft_m}:=\hat{\bft}_m-\bar{\bft}$. 
\end{itemize}
Fig.~\ref{fig:compareellipsesyn} shows projections of the covariance
ellipsoids on pairs of axes for the Monte-Carlo method and our method,
when $\lambda = 10^{-5}$. It can be noted that the proposed method
provides an excellent estimation of the covariance of the hand-eye
transformation.

\begin{figure}[t]
  \centering
  \includegraphics[width=0.4\textwidth]{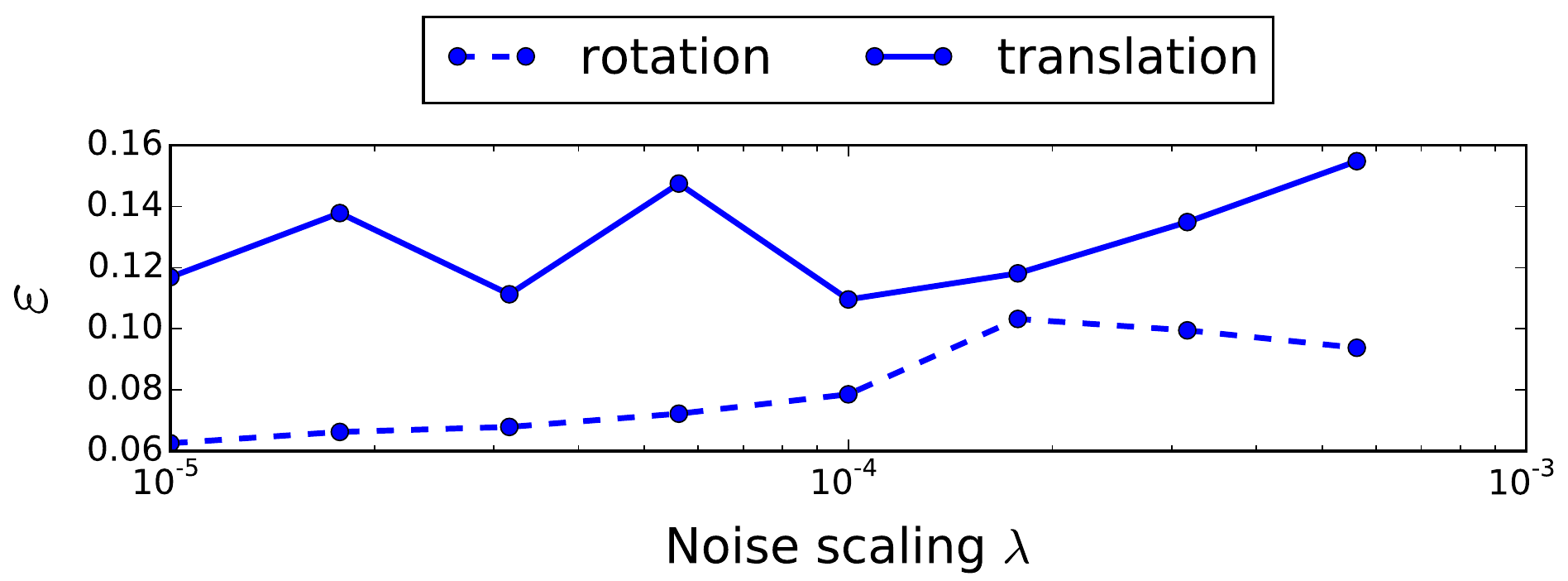}
  \caption{Difference between the covariances obtained from
    Monte-Carlo and predicted by our method, at different noise
    levels.}
  \label{fig:comparefrobenius}
\end{figure}

To gauge the performance at different noise levels, we use the
following metrics
{\small
\begin{equation}
\varepsilon = \frac{\sqrt{\text{tr}((\bfSigma^\mathrm{PR}-\bfSigma^\mathrm{MC})^\top(\bfSigma^\mathrm{PR}-\bfSigma^\mathrm{MC}))}}{\sqrt{\text{tr}({\bfSigma^\mathrm{MC}}^\top \bfSigma^\mathrm{MC})}}.
\end{equation}}
Fig~\ref{fig:comparefrobenius} shows that our algorithm can cope well
with increasing magnitudes of the measurement uncertainty. The
estimation errors remain low overall, and increases slightly with the
magnitude of the noise, since larger noise levels increase the number
of local minima at each iteration. Note also that the errors in the
covariances of the translation parts tend to be larger than that of
the rotation parts. This is because, in our method, the errors in the
estimation of the rotation propagate to that of the
translation. Regarding the computation cost, our method is
naturally several magnitude faster than the Monte-Carlo method.

It is also worth noting that the closed form solution
in~\cite{FM94tra} always yields slightly higher covariance as compared
to our method. This is because our method does optimally minimize the
error of the estimated transformation by taking to account the
measurement noise.

\subsection{Real calibration data}
\label{sec:realcalibdata}
We now validate the proposed method on actual calibration data
obtained from our robot system, which consists of a 3D camera mounted
on a 6-DOF industrial manipulator, as shown in
Fig.~\ref{fig:method}.

\subsubsection{Covariances of \texorpdfstring{$\bmA$}{A} and \texorpdfstring{$\bmB$}{B} in the actual system}

We first need to empirically estimate the covariances of the
$\bmA_i$'s and $\bmB_i$'s in our system, so that we can give them as
inputs to our method. 

As the industrial manipulator has a very high precision (0.2 mm of
repeatability), we assume that the noise on the $\bmA_{i}$'s is
negligible.

Regarding the $\bmB_{i}$'s, the $\bmB_1,\dots,\bmB_k$ are assumed to have the same noise distributions: $\forall i,\
\bfSigma_{\bmR_{\bmB_i}}=\bfSigma_{\bmR_{\bmB}},
\bfSigma_{\bft_{\bmB_i}}=\bfSigma_{\bft_{\bmB}}$. We
experimentally collect 500 pairs of $\bmA_i$ and $\bmB_i$ from our
system. Next, we generate M = 400 datasets, each dataset comprising $k =
30$ pairs $(\bmA_i , \bmB_i)$ randomly selected from the collected
pairs. 

The rotation and translation errors of $\bmB_i$'s are then computed as
{\small
\begin{eqnarray}
\label{eq:errorrotreal}
\bfxi_{\bmR_{\bmB_i}}&=&(\log(\bmR_{\bmB_i}\bar{\bmR}_{\bmB_i}^{-1}))^{\vee},\\
\bfxi_{\bft_{\bmB_i}}&=&\bft_{\bmB_i}-\bar{\bft}_{\bmB_i},\label{eq:errortransreal}
\end{eqnarray}} where the ground truth is $\bar{\bmB}_{i} =
{\bar{\bmX}}^{-1}{\bmA}_{i}\bar{\bmX}$.

Since the true
transformation $\bar{\bmX}$ is unknown in the real system, we use 
{\small
\begin{eqnarray}
\label{eq:xrotavg}
\bmR^\mathrm{avg}&=&\exp{\left(\left[\frac{1}{M}\sum_{m}^{M}(\log{\hat{\bmR}_{m}})^{\vee}\right]\right)},\\
\label{eq:xtransavg} \bft^\mathrm{avg}&=&\frac{1}{M}\sum_{m}^{M}\hat{\bft}_{m},
\end{eqnarray}}
as the ground truth. Note that estimating $\hat{\bmR}_{m},
\hat{\bft}_{m}$ using our method would require information of
$\bmA_i$'s, $\bmB_i$'s noise, therefore we use~\cite{FM94tra} instead.

After obtaining the rotation and translation errors of $\bmB_i$'s, the
empirical covariance matrices of $\bmB_i$ can be estimated similarly
to the equations (\ref{eq:mc_rot},\ref{eq:mc_trans}).

\subsubsection{Validation}

To validate our method, we collect another 500 pairs of $\bmA_{i}$ and
$\bmB_{i}$ from our system. We constrain the robot motion so that it
covers the same area as that used for determining the noise on
$\bmB_i$. We then generate $M=400$ datasets, each dataset comprising
$k=30$ pairs $(\bmA_{i},\bmB_{i})$ randomly selected from the
collected pairs. The covariance matrices are computed from these
datasets using the Monte-Carlo method and our method, in the same
manner as previously.

\begin{figure}[t]
  \centering
  \includegraphics[width=0.4\textwidth]{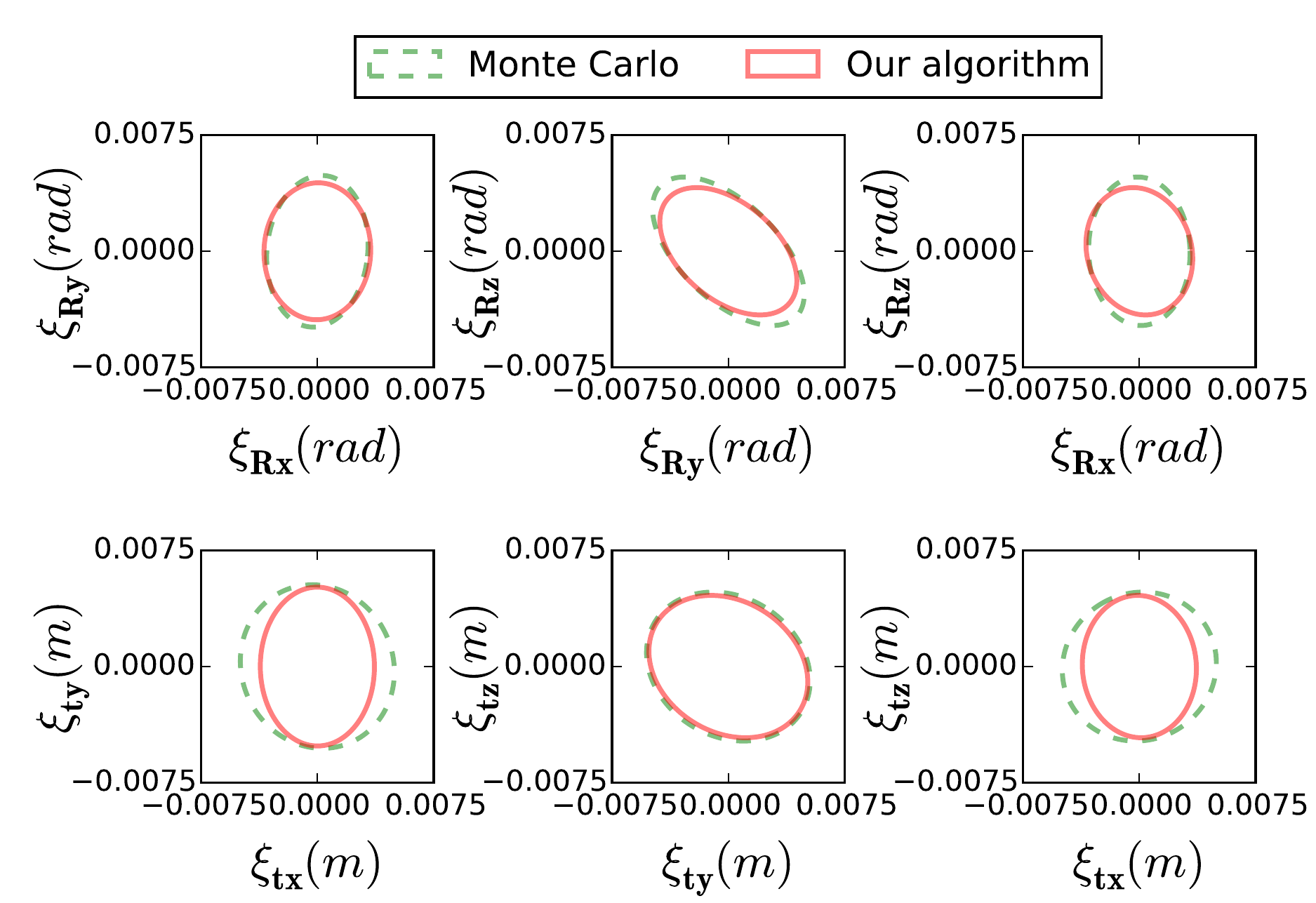}
  \caption{Projection of the one-standard-deviation covariance
    ellipsoids on pairs of axes for $\bfSigma_{\bmR}$ and
    $\bfSigma_{\bft}$, shown for Monte-Carlo and our algorithm with
    real data, axes' labels are similar as to Fig.~\ref{fig:compareellipsesyn}.}
  \label{fig:compareellipsereal}
\end{figure}
Fig.~\ref{fig:compareellipsereal} provides projections of the covariance
ellipsoids on
pairs of axes shown for two methods. We see that the proposed method
delivers a good estimation of the covariances. We do not believe
there has been another methods of estimating uncertainty of the
hand-eye transformation. Moreover, the proposed method is also relatively
easy to replicate and use in practical applications.

\subsection{Covariance of the object pose estimation}

Using the covariance of $\bmX$ previously obtained, we are now in a
position to predict the covariance of the object pose estimation,
which is our ultimate goal. Here, we demonstrate the propagation of
uncertainties to the object pose estimation using the same robot
system as previously (see Fig.~\ref{fig:method}).


Recall that the constant transformation of the pattern (object) with
respect to the robot base is given by
{\small
\begin{equation}
\label{eq:yseries}
\bmY \coloneqq \tf{\bfb}{\bmT}{\bfo} = \tf{\bfb}{\bmT}{\bfe}_{i}\tf{\bfe}{\bmT}{\bfc}\tf{\bfc}{\bmT}{\bfo}_{i}.
\end{equation}}
The covariances of $\tf{\bfb}{\bmT}{\bfe}_{i}$ and
$\tf{\bfe}{\bmT}{\bfc}$ $(=\bmX)$ can be estimated using the procedure
proposed in Section~\ref{sec:realcalibdata}. Thus, to predict the mean
and the covariance of $\bmY$, one needs now to estimate the
covariances of $\tf{\bfc}{\bmT}{\bfo}_{i}$.

Suppose that $\tf{\bfc}{\bmT}{\bfo}_{1},...,\tf{\bfc}{\bmT}{\bfo}_{k}$
have the same noise distribution,~\ie{} $\forall
i,\ \bfSigma_{\bmR_{\tf{\bfc}{\bmT}{\bfo}
    i}}=\bfSigma_{\bmR_{\bfc\bfo}},
\bfSigma_{\bft_{\tf{\bfc}{\bmT}{\bfo} i}}=\bfSigma_{\bft_{\bfc\bfo}}$.
We begin by experimentally collecting 500 pairs of
$\tf{\bfb}{\bmT}{\bfe}_{i}, \tf{\bfc}{\bmT}{\bfo}_{i}$ from our
system. The rotation and translation errors of
$\tf{\bfc}{\bmT}{\bfo}_{i}$’s are then computed similar to
(\ref{eq:errorrotreal}, \ref{eq:errortransreal}), where the ground
truth is $\bar{\tf{\bfc}{\bmT}{\bfo}_{i}} = \bar{\bmY}
\bar{\tf{\bfb}{\bmT}{\bfe}_{i}}\bar{\bmX}$. As discussed earlier, we will use (\ref{eq:xrotavg},\ref{eq:xtransavg})
instead of the true value of $\bar{\bmX}$. Regarding $\bar{\bmY}$, one can
transform~(\ref{eq:yseries}) into
{\small
\begin{equation}
{\tf{\bfb}{\bmT}{\bfe}_{p}} {\tf{\bfb}{\bmT}{\bfe}_{q}}^{-1}
\bmY= \bmY
{\tf{\bfc}{\bmT}{\bfo}_{p}}^{-1} {\tf{\bfc}{\bmT}{\bfo}_{q}}, p
\neq q,
\end{equation}}
which has the form of $\bmA^\prime_i \bmY = \bmY\bmB^\prime_i$, where
$\bmA^\prime\coloneqq {\tf{\bfb}{\bmT}{\bfe}_{p}}
{\tf{\bfb}{\bmT}{\bfe}_{q}}^{-1}$ and $\bmB^\prime \coloneqq
{\tf{\bfc}{\bmT}{\bfo}_{p}}^{-1} {\tf{\bfc}{\bmT}{\bfo}_{q}}$. Hence,
$\bar{\bmY}$ can also be computed in the same manner as
computing $\bar{\bmX}$.

We now collect 500 pairs of $\tf{\bfb}{\bmT}{\bfe}_{i},
\tf{\bfc}{\bmT}{\bfo}_{i}$ from our system. We then generate $M = 400$
datasets, each dataset comprising $k = 30$ $(\bmA_i,\bmB_i)$ pairs
computed from $(\tf{\bfb}{\bmT}{\bfe}_{i},
\tf{\bfc}{\bmT}{\bfo}_{i})$ pairs randomly selected from the 500 collected
pairs.

Next, we evaluate the covariance of $\bmY$ following two methods
\begin{itemize}
\item \emph{Our method:} For some dataset $m\in[1,M]$, we compute the
  covariance matrices $\bfSigma_{\bmR_{\bmY m}},\bfSigma_{\bft_{\bmY
      m}}$ using the propagation method described in Appendix
  \ref{sec:compoundingtheory} (\underline{PR}edicted). In fact, the
  $\bfSigma_{\bmR_{\bmY m}},\bfSigma_{\bft_{\bmY m}}$ are nearly identical across
  the $M$ datasets, so the particular value of $m$ did not matter;
\item \emph{Monte-Carlo}: For each dataset $m \in [1, M ]$, we compute
  $\bmY_m = \tf{\bfb}{\bmT}{\bfo}_m= \tf{\bfb}{\bm
  \mathring{T}}{\bfe}\hat{\tf{\bfe}{\bmT}{\bfc}_m}\tf{\bfc}{\bm
  \mathring{T}}{\bfo}$, where $\tf{\bfb}{\bm \mathring{T}}{\bfe}$ and
  $\tf{\bfc}{\bm \mathring{T}}{\bfo}$ are randomly selected from the
  collected pairs; the rotation and translation of
  $\hat{\tf{\bfe}{\bmT}{\bfc}_m}$ together with their covariances are
  computed using our method as proposed in
  Section~\ref{sec:derivation} (see
  equations~(\ref{eq:sigmarotation},~\ref{eq:sigmatranslation})).
  Next, the covariances are computed by the
  \underline{M}onte-\underline{C}arlo method as $
  \bfSigma_{\bmR_{\bmY}}^\mathrm{MC}:=\frac{1}{M}\sum_{m}^{M}{\bfxi_{\bmR_{\bmY_m}}}
  \bfxi^\top_{\bmR_{\bmY_m}}$ and
  $\bfSigma_{\bft_{\bmY}}^\mathrm{MC}:=\frac{1}{M}\sum_{m}^{M}\bfxi_{\bft_{\bmY_m}}
       {\bfxi^\top_{\bft_{\bmY_m}}},$ where
       $\bfxi_{\bmR_{\bmY_m}}:=(\log(\hat{\bmR}_{\bmY_m}
       {\bar{\bmR}_{\bmY}}^{-1}))^{\vee}$,
       $\bfxi_{\bft_{\bmY_m}}:=\hat{\bft}_{\bmY_m}-\bar{\bft}_{\bmY}$.
\end{itemize}

\begin{figure}[t]
  \centering
  \includegraphics[width=0.4\textwidth]{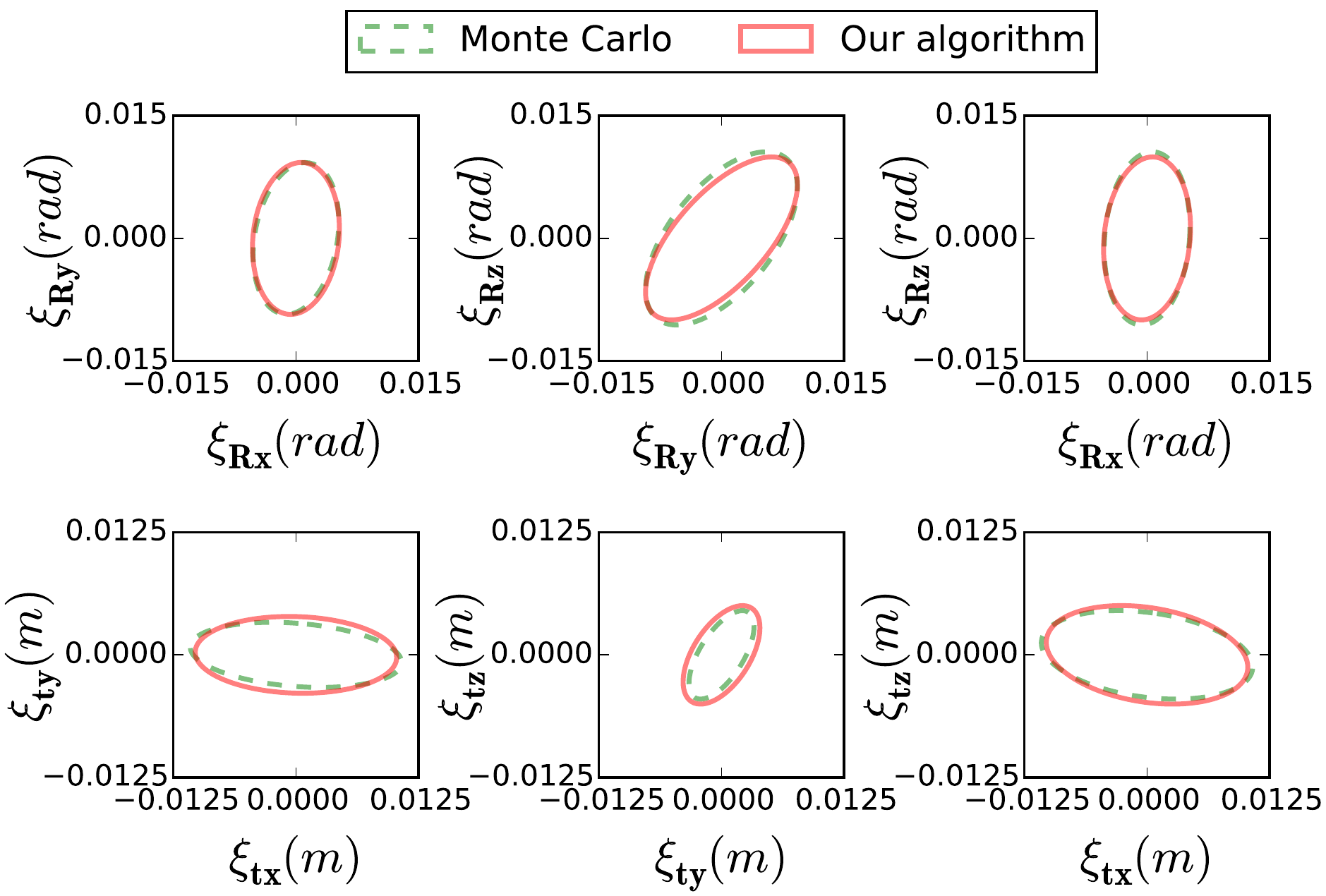}
  \caption{Projection of the one-standard-deviation covariance
    ellipsoids on pairs of axes for $\bfSigma_{\bmR_{\bmY}}$ and
    $\bfSigma_{\bft_{\bmY}}$, shown for Monte-Carlo and our algorithm,
    axes' labels are similar as to Fig.~\ref{fig:compareellipsesyn}.}
  \label{fig:propagate-result}
\end{figure}

Fig.~\ref{fig:propagate-result} shows the one-standard-deviation
covariance ellipsoids shown for two methods. One can see that our
prediction matches very well the covariances estimated by the
Monte-Carlo method. 

In absolute values, the covariance of the hand-eye calibration
compounds with that of the object pose estimation in the camera frame,
resulting in a relatively large overall covariance for the object pose
estimation in the robot frame, around 1cm in standard deviation. This
again emphasizes the need of having access to the covariance of the
hand-eye transformation. This fine-grained information tells us how
confident we can be regarding the object pose estimation and shall
also enable us to design new perception algorithms and methods for
reaching higher precision, by \eg{} visuo-tactile sensor fusion.

\section{Conclusion}
\label{sec:conclusion}
In this paper, we have presented a rigorous derivation of the
covariance of the solution $\bmX$, when $\bmA$ and $\bmB$ are randomly
perturbed matrices. Our approach consists in transposing methods for
forward and backward propagation of covariance into the framework of
uncertainty in $\SE$. Experiments involving synthetic and real
calibration data show that our approach can predict the covariance of
the hand-eye transformation with excellent precision.

While these estimates could also be provided by Monte-Carlo
simulations, such a method would require collecting a large number of
samples, which is not practical. Furthermore, the Monte-Carlo method
yields no insights into how the uncertainties on the measurements of
$\bmA$ and $\bmB$ propagate to the uncertainty of the hand-eye
transformation. By contrast, in our method, by analyzing critical
factors influencing the covariance of $\bmX$, for instance, based on
the formulae~(\ref{eq:sigmarotation}) and~(\ref{eq:sigmatranslation}),
one may be able to refine the calibration process to achieve a higher
precision, by \eg{} determining the appropriate number of sample
viewpoints or choosing their optimal distribution, which is the object
of our future research.

\section*{Acknowledgment}
This work was partially supported by NTUitive Gap Fund NGF-2016-01-028.

\appendices
\section{Propagating uncertainties when rotation and translation
are decoupled}
\label{sec:compoundingtheory}
In this Section, we present our extension of the covariance
propagation method of \cite{Barfoot14tr} to the case where rotation
and translation are decoupled.

Consider two noisy poses $\bmT_1$ and $\bmT_2$, whose
nominal values and associated uncertainties are $ \{
{\bar{\bmR}}_{1},\bfSigma_{\bmR_{1}} \}, \{
{\bar{\bft}}_{1},\bfSigma_{\bft_{1}} \}$ and $
\{{\bar{\bmR}}_{2},\bfSigma_{\bmR_{2}}\}, \{
  {\bar{\bft}}_2,\bfSigma_{\bft 2} \}$ respectively.

Let $\bmT_{12}
    = \bmT_1\bmT_2$ be the compounded pose, we have
{\small
\begin{eqnarray}
\label{eq:compoundR}
 {\bar{\bmR}}_{12}&=& \bar{\bmR}_1\bar{\bmR}_2,\\
\label{eq:compoundt}
 {\bar{\bft}}_{12}&=&  \bar{\bmR}_1\bar{\bft}_2 +\bar{\bft}_1.
\end{eqnarray}}
Similar to \cite{Barfoot14tr} (Section III), the covariance
matrix of the rotation can be estimated by:  
{\small
\begin{eqnarray}
\lefteqn{\bfSigma_{\bmR_{12}} \approx\bfSigma_{\bmR_{1}} +
  \bfSigma^{\prime}_{\bmR_{2}}} \nonumber\\
&+\frac{1}{12} ({\bm{\calA}}_1\bfSigma^{\prime}_{\bmR_{2}} +
  \bfSigma^{\prime}_{\bmR_{2}} {{\bm{\calA}}_1}^\top+ \bfSigma_{\bmR_{1}}
    {\bm{\calA}}_2 + \bfSigma_{\bmR_{1}} {{\bm{\calA}}_2}^\top) +
    \frac{\bm{\calB}}{4}, \label{eq:compoundcovR}\\
\text{where}& \bfSigma^{\prime}_{\bmR_{2}} := \bar{\bmR}_1\bfSigma_{\bmR_{2}}{\bar{\bmR}_1}^\top,\\
&{\bm{\calA}}_1:=\llangle \bfSigma_{\bmR_{1}} \rrangle,
{\bm{\calA}}_2:=\llangle \bfSigma^{\prime}_{\bmR_{2}} \rrangle,
\bm{\calB}:=\llangle \bfSigma_{\bmR_{1}},\bfSigma^{\prime}_{\bmR_{2}}
\rrangle,\\
\text{and}& \llangle \bmM \rrangle
\coloneqq -\textrm{tr}(\bmM) \bbI + \bmM,\\
&\llangle \bmM,\bmN \rrangle
\coloneqq \llangle\bmM\rrangle
\llangle\bmN\rrangle+\llangle\bmN\bmM\rrangle,
\end{eqnarray}} with {\small$\bmM,\bmN \in \bbR^{n \times n}$.}

Regarding the translation vector, its covariance matrix can be
estimated simply by using the forward propagation method of
Section~\ref{sec:fwdbwdpropagation}:
{\small
\begin{equation}
\label{eq:compoundcovt}
\bfSigma_{\bft_{12}} \approx \bfSigma_{\bft_{1}} +
\bar{\bmR}_{1}\bfSigma_{\bft_{2}}{\bar{\bmR}_{1}}^\top+ [{\bar{\bmR}_{1}}{\bar{\bft}_{2}}]\bfSigma_{\bmR_{1}}[{\bar{\bmR}_{1}}{\bar{\bft}_{2}}]^\top.
\end{equation}}
In summary, to compound two poses, we propagate the means
using (\ref{eq:compoundR},\ref{eq:compoundt}) and the covariances
using (\ref{eq:compoundcovR},\ref{eq:compoundcovt}).

\bibliographystyle{IEEEtran}
\bibliography{axxb.bbl}
\end{document}